\documentclass[11pt,a4paper]{article}
\usepackage[T1]{fontenc}
\usepackage[utf8]{inputenc}
\usepackage{lmodern}
\usepackage[margin=0.91in,headheight=14pt]{geometry}
\usepackage{amsmath,amssymb,mathtools}
\usepackage{graphicx,booktabs,tabularx,array}
\usepackage{microtype}
\usepackage{xcolor}
\usepackage{enumitem}
\usepackage{caption}
\usepackage{fancyhdr}
\usepackage{hyperref}
\usepackage{placeins}
\usepackage{tikz}
\usetikzlibrary{arrows.meta,positioning,calc,fit}
\definecolor{ink}{HTML}{17334D}
\definecolor{linkblue}{HTML}{225E8B}
\hypersetup{colorlinks=true,linkcolor=linkblue,citecolor=linkblue,urlcolor=linkblue,pdftitle={WildMAE: Structure-Aware Masked Autoencoding for Wildlife Image Reconstruction},pdfauthor={Yicheng Zhang, Yujie Jin, Ke Zhang, Jiaxin Wu},pdfsubject={Research preprint}}
\usepackage{titlesec}
\titleformat{\section}{\large\bfseries\color{ink}}{\thesection}{0.7em}{}
\titleformat{\subsection}{\normalsize\bfseries\color{ink}}{\thesubsection}{0.65em}{}
\titlespacing*{\section}{0pt}{14pt}{6pt}
\titlespacing*{\subsection}{0pt}{10pt}{4pt}
\setlist[itemize]{leftmargin=1.35em,topsep=4pt,itemsep=2pt,parsep=0pt}
\newcommand{\Lpix}{\mathcal{L}_{\mathrm{pix}}}
\newcommand{\Ledge}{\mathcal{L}_{\mathrm{edge}}}
\newcommand{\Lperc}{\mathcal{L}_{\mathrm{perc}}}
\newcommand{\mset}{\mathcal{M}}
\newcommand{\vset}{\mathcal{V}}

\title{\vspace{-1.2em}\textbf{\color{ink}WildMAE: Structure-Aware Masked Autoencoding\\for Wildlife Image Reconstruction}}
\author{Yicheng Zhang \qquad Yujie Jin \qquad Ke Zhang \qquad Jiaxin Wu\\[0.3em]
\normalsize The Chinese University of Hong Kong, Shenzhen}
\date{}
\begin{document}
\maketitle\vspace{-1.1em}
\begin{abstract}
\noindent Camera-trap imagery presents substantial variation in animal appearance, illumination, background clutter, and visual degradation. Masked autoencoders (MAEs) offer a label-free framework for modeling such data through reconstruction of randomly hidden image patches. We present \textbf{WildMAE}, a decoder-side extension of MAE that combines local structure-aware refinement with masked-to-visible cross-attention. A zero-gated convolutional residual path introduces a two-dimensional neighborhood prior, while gated cross-attention allows masked positions to retrieve contextual information from visible-position features without replacing the original Transformer decoder. The associated training recipe progressively introduces normalized-pixel, Sobel-edge, and perceptual objectives. In the original Snapshot Serengeti study, covering 4,744 test images at 75\% masking, the reported reconstruction scores were 27.90 versus 24.33 dB PSNR, 0.840 versus 0.780 SSIM, and 0.160 versus 0.320 LPIPS for WildMAE and vanilla MAE, respectively. Additional historical records document favorable average differences under four synthetic degradation conditions and describe component variants. We present the method, experimental observations, and their scope together: the retained metrics are historical protocol-dependent results, rather than a newly reproduced, harmonized comparison or a validation of restoration from genuinely unavailable pixels. WildMAE provides a practical architectural case study for combining global token mixing with spatial and targeted contextual refinement in wildlife masked-image modeling.
\end{abstract}
\vspace{0.3em}\noindent\textbf{Keywords:} masked autoencoders; wildlife camera traps; image reconstruction; cross-attention; structural refinement.

\section{Introduction}
Camera traps make it possible to observe wildlife at scale without continuous human presence, but the resulting images vary substantially in illumination, background, object scale, occlusion, and motion. In resources such as Snapshot Serengeti~\cite{swanson2015snapshot}, even adjacent frames can depict challenging combinations of animals and cluttered natural scenes. This variability motivates image modeling approaches that can exploit visual structure without relying exclusively on species or behavior annotations. Masked image modeling is particularly appealing: a network receives a sparse set of visible patches and learns to predict those withheld during training.

Masked autoencoders (MAEs)~\cite{he2022mae} use an asymmetric Vision Transformer (ViT)~\cite{dosovitskiy2021vit} encoder--decoder, processing only visible patches in the encoder and reconstructing masked targets with a lightweight decoder. The full-sequence decoder already integrates positional embeddings and global self-attention; however, its patch sequence need not explicitly prioritize local two-dimensional adjacency after token mixing. For wildlife imagery, spatial continuity and targeted access to the remaining scene may be useful when predicting silhouettes, contours, and textured regions from sparse observations. This motivates an additional decoder-side inductive bias rather than a replacement of the MAE pathway.

\begin{figure}[t]
\centering
\includegraphics[width=0.88\linewidth]{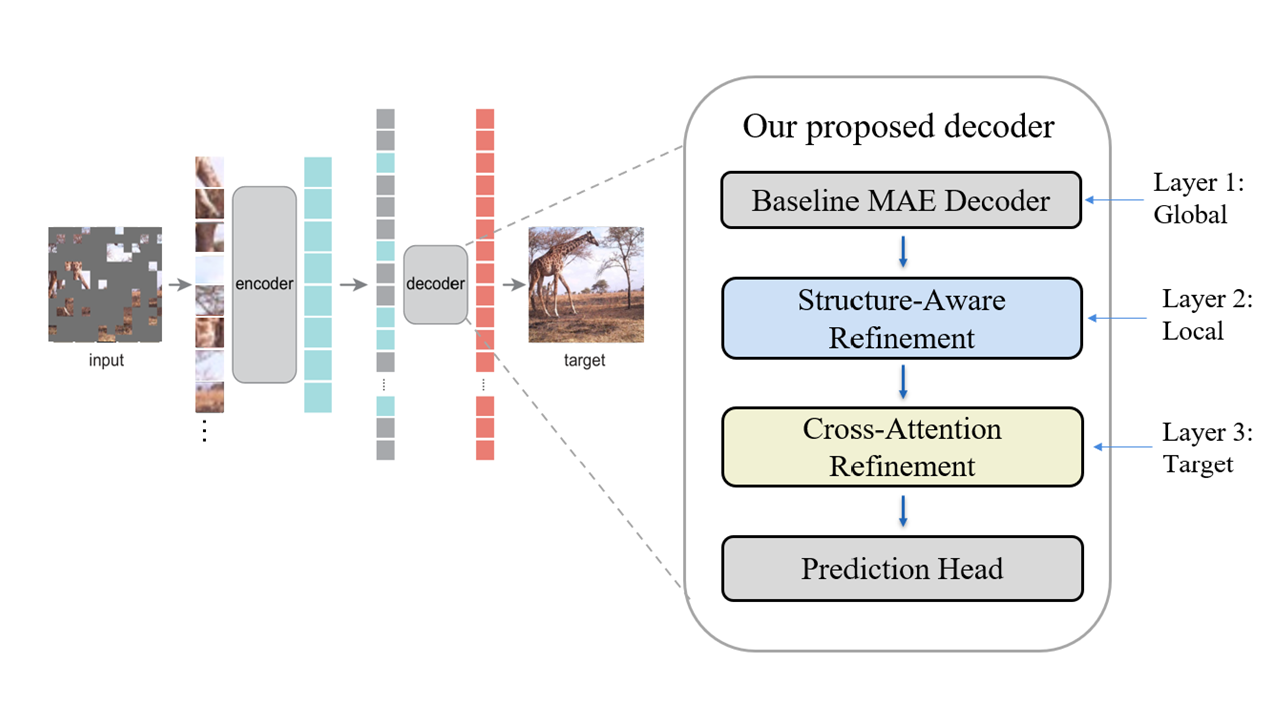}
\caption{High-level WildMAE pipeline: the original MAE encoder and Transformer decoder are retained, followed by a local spatial refinement, masked-to-visible contextual refinement, and the pixel-prediction head. Exact internal normalizations and residual gates are specified in Sections~\ref{sec:local}--\ref{sec:cross}. The diagram is conceptual.}
\label{fig:overall}
\end{figure}

We study \textbf{WildMAE}, which retains the global MAE decoder and appends two complementary refinements (Figure~\ref{fig:overall}). First, a zero-gated local convolutional block processes decoder patch tokens on their restored spatial grid. Second, masked-position queries attend directly to features at visible positions, followed by a separately gated feed-forward update. In the accompanying training design, pixel reconstruction is established before edge and perceptual supervision are introduced. The intent is to combine the global context of self-attention, the locality of spatial convolution, and information retrieval targeted at masked regions.

The original project investigated this design on a 47,349-image checked subset of Snapshot Serengeti, reporting a 75\%-masking comparison, component variants, and evaluations with synthetic degradations. The present manuscript turns that study into a self-contained method and results account while retaining the original measurements and specifying the evaluation setting more precisely. In particular, it distinguishes normalized-patch prediction from RGB reconstruction that uses full-input patch statistics, an important distinction when interpreting the image-quality scores.

The contributions of this work are:
\begin{itemize}
\item A decoder refinement architecture combining a local spatial residual branch and masked-to-visible cross-attention while preserving the original MAE decoder pathway.
\item An explicit description of the zero-gated refinement updates and a progressive pixel--edge--perceptual training formulation.
\item An organized presentation of the original wildlife-image reconstruction, component, and synthetic-degradation observations, together with clear boundaries on their interpretation and reproducibility.
\end{itemize}

\section{Related Work}
\paragraph{Wildlife camera-trap vision.} Snapshot Serengeti provides a large, annotated collection of camera-trap photographs~\cite{swanson2015snapshot}. Related work has addressed animal identification, counting, and description~\cite{norouzzadeh2018automatically}, as well as filtering empty frames~\cite{cunha2021filtering}. Such recognition pipelines and masked reconstruction address different questions: reconstructing withheld image regions is an image-modeling task, not direct evidence of improved ecological recognition or a substitute for observations of real animals.

\paragraph{Masked image modeling.} ViTs model image patches as tokens~\cite{dosovitskiy2021vit}, and MAE reconstructs randomly masked patches from visible context~\cite{he2022mae}. CrossMAE studies patch dependence using masked queries and visible features for efficient decoding~\cite{fu2025crossmae}. WildMAE differs in retaining the MAE full-token self-attention decoder and adding a separate masked-query refinement afterward; consequently, its design should not be interpreted as inheriting CrossMAE's computational-efficiency claims. MCMAE~\cite{gao2022mcmae} combines convolution and masked autoencoding, underscoring that local processing is not itself a novel ingredient. The distinction here is the placement of a local spatial branch \emph{after} global decoder processing, followed by visible-to-masked feature retrieval. HAP~\cite{yuan2023hap} introduces human-part-based structure priors; WildMAE instead uses a generic patch-grid prior without part labels.

\paragraph{Training objectives and perceptual quality.} Pixel reconstruction measures target fidelity, whereas Sobel gradients emphasize local image boundaries. SSIM~\cite{wang2004ssim} and learned perceptual image patch similarity (LPIPS)~\cite{zhang2018lpips} characterize complementary aspects of visual agreement. Progressive supervision in WildMAE changes the terms of the reconstruction objective; it is distinct from learning a masking curriculum as in CL-MAE~\cite{madan2024clmae}. Because LPIPS is also a training component in the final stage, LPIPS evaluation is best interpreted alongside, rather than instead of, pixel and structural measurements.

\section{Proposed Method}
\subsection{Overview and MAE backbone}
Let $I\in\mathbb{R}^{224\times224\times3}$ be an RGB image. Patch size $16\times16$ yields $N=14\times14=196$ patches of dimension $D=16^2\cdot3=768$. For a masking ratio $r=0.75$, the index sets of visible patches $\vset$ and masked patches $\mset$ have sizes 49 and 147. The retained ViT-B encoder contains 12 blocks with width 768 and 12 attention heads. It processes visible tokens, after which learned mask tokens and positional embeddings restore a full-sequence decoder input. The decoder provides a global context representation that the proposed refinements further process.

As in normalized-pixel MAE training, let $\mathbf p_i$ denote a channel-normalized input patch and write its per-patch mean and variance as $\mu_i$ and $s_i^2$. The prediction target and masked-region pixel objective are
\begin{equation}
\widetilde{\mathbf p}_i=\frac{\mathbf p_i-\mu_i}{\sqrt{s_i^2+\epsilon}},\qquad
\Lpix=\frac{1}{|\mset|D}\sum_{i\in\mset}\bigl\|\widehat{\mathbf p}_i-\widetilde{\mathbf p}_i\bigr\|_2^2,
\quad\epsilon=10^{-6}.
\label{eq:pixel}
\end{equation}
The normalization is used to construct the supervision target. Interpretation of the separate RGB reconstruction step is discussed in Section~\ref{sec:protocol}.

\subsection{Structure-aware local refinement}\label{sec:local}
Let $X\in\mathbb{R}^{N\times C}$ be the restored patch-token sequence after the MAE decoder blocks, excluding the class token. A reshape operator $\mathcal R$ maps patch tokens to their $C\times14\times14$ grid. The local branch applies pre-normalization, a depthwise $3\times3$ convolution, and two pointwise projections with a GELU nonlinearity:
\begin{align}
F(X)&=\mathcal R^{-1}\!\left[P_2\!\left(\operatorname{GELU}\!\left(P_1\!\left(\operatorname{DWConv}_{3\times3}(\mathcal R(\operatorname{LN}(X)))\right)\right)\right)\right],\label{eq:local}\\
S&=X+\gamma_sF(X).\label{eq:localgate}
\end{align}
The depthwise convolution uses padding one; the pointwise projections expand from $C$ to $2C$ and return to $C$. The learnable scalar gate $\gamma_s$ is initialized to zero, so the additional local branch initially preserves its input. This property refers to the added residual branch, not to equivalence between differently sized complete decoders. Explicitly restoring the two-dimensional token layout encourages local neighborhood interactions while preserving information already combined by global self-attention.

\subsection{Masked-to-visible cross-attention}\label{sec:cross}
After local refinement, use the restored mask to gather masked-token features $S_{\mset}$ and visible-position features $S_{\vset}$. Define $U^{(0)}=S_{\mset}$ and $Z=S_{\vset}$. At refinement block $\ell$, masked tokens form queries and visible-position tokens form keys and values:
\begin{align}
A^{(\ell)} &= \operatorname{MHA}\left(\operatorname{LN}_q(U^{(\ell)}),\operatorname{LN}_{kv}(Z),\operatorname{LN}_{kv}(Z)\right),\label{eq:attn}\\
T^{(\ell)} &= U^{(\ell)}+\gamma_a^{(\ell)}A^{(\ell)},\label{eq:attngate}\\
U^{(\ell+1)} &= T^{(\ell)}+\gamma_f^{(\ell)}\operatorname{MLP}\bigl(\operatorname{LN}_f(T^{(\ell)})\bigr).\label{eq:mlpgate}
\end{align}
The attention operation includes learned query/key/value/output projections, and the feed-forward network has a fourfold hidden-width expansion with GELU. The attention and MLP residual gates, $\gamma_a^{(\ell)}$ and $\gamma_f^{(\ell)}$, are independently initialized to zero. The retained constructors define two refinement blocks with eight heads, though the surviving files do not identify every historical checkpoint with equal precision. Visible-position features stay fixed within this stack, while the updated masked features are scattered back to their original token indices. After reattaching the class token, final decoder normalization and a prediction head produce normalized patch values.

\begin{figure}[t]
\centering
\includegraphics[width=0.90\linewidth]{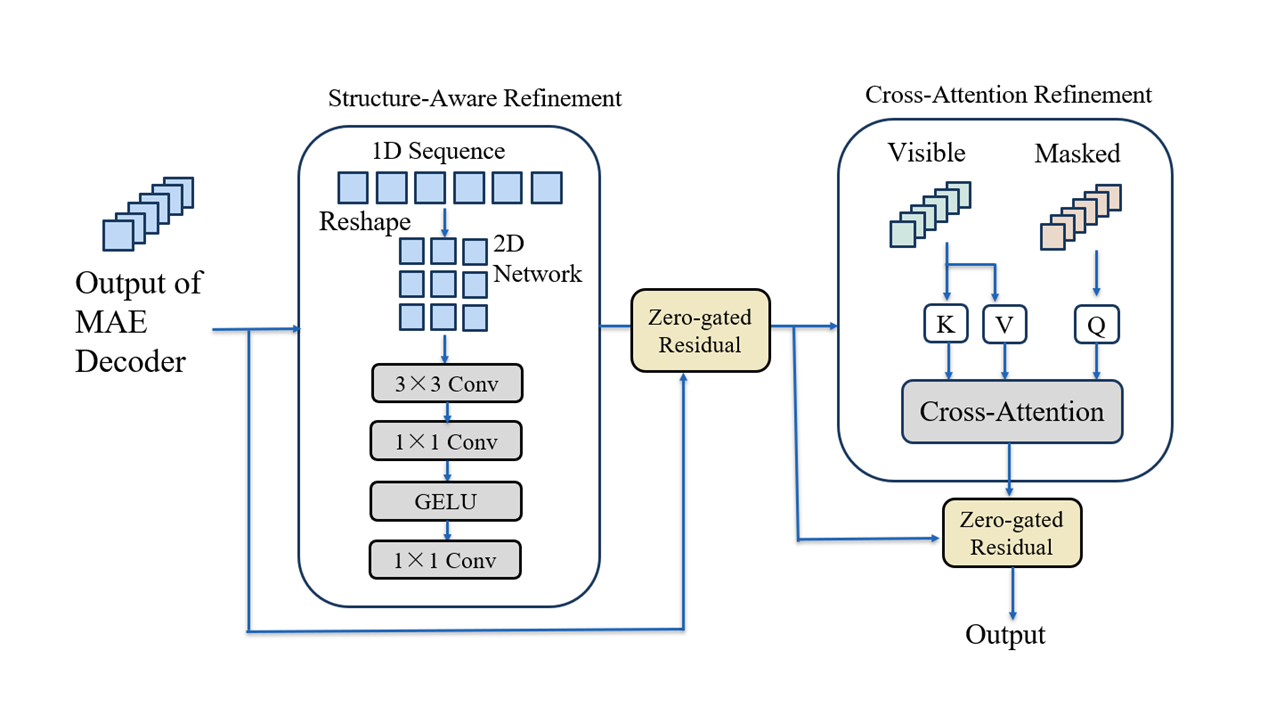}
\caption{Conceptual illustration of the two appended refinement operations, adapted from the original project report. The actual implementation uses depthwise convolution and pre-normalization in the local branch, plus separately zero-gated cross-attention and MLP updates in each cross-refinement block (Equations~\ref{eq:local}--\ref{eq:mlpgate}).}
\label{fig:modules}
\end{figure}

Notably, $Z$ comes from the globally mixed and locally refined decoder representation: it is \emph{not} a set of untouched raw image patches. The local and targeted branches therefore complement, rather than substitute for, full-token decoder processing. Figure~\ref{fig:modules} summarizes the conceptual information flow.

\subsection{Progressive structural--perceptual supervision}
The original training plan has three phases. Stage 1 optimizes $\Lpix$ for 160 epochs. Stage 2 adds edge supervision for 20 epochs. Stage 3 includes both the edge term and an LPIPS term for another 20 epochs:
\begin{equation}
\mathcal L=\Lpix+\lambda_e\Ledge+\lambda_p\Lperc,
\label{eq:objective}
\end{equation}
with planned coefficients $(\lambda_e,\lambda_p)=(0,0)$, $(0.02,0)$, and $(0.02,0.002)$ in the respective phases. These are the weights specified by the staged pipeline; other generic scripts have different defaults and are not evidence that those defaults produced the main reported checkpoint.

For edge and perceptual losses, retained fine-tuning code converts predicted normalized patches into RGB and creates a composite $\widehat I$ from visible input pixels and predicted masked pixels. With pixel mask $m$ equal to one in masked regions, the Sobel magnitude of the channel-averaged image is $G(I)$ and the masked edge term is
\begin{equation}
\Ledge=\frac{\sum m\,\bigl|G(\widehat I)-G(I)\bigr|}{\sum m}.
\label{eq:edge}
\end{equation}
The perceptual term replaces visible pixels with gray in both inputs before measuring AlexNet-based LPIPS:
\begin{equation}
C_m(J)=m\odot J+(1-m)\odot0.5,\qquad
\Lperc=\operatorname{LPIPS}_{\mathrm{AlexNet}}\bigl(2C_m(\widehat I)-1,2C_m(I)-1\bigr).
\label{eq:perceptual}
\end{equation}
The perceptual network is frozen during fine-tuning; gradients with respect to its reconstructed-image input remain available. This staged design balances pixel agreement with progressively introduced structural and perceptual constraints, without requiring species-class labels as training targets.

\section{Experimental Setup}
\subsection{Dataset, preprocessing, and training}
The study planned a 50,000-image subset of Snapshot Serengeti Season 1 and retained 47,349 checked image IDs, comprising 37,903 training, 4,702 validation, and 4,744 test records. Input images are center-cropped to a square and resized to $224\times224$ for evaluation; pretraining additionally uses random resized crops and horizontal flips. The checked subset includes animal and empty-frame images. Training and the primary reconstruction study use a fixed 75\% random patch-masking ratio.

The reported optimizer is AdamW with weight decay 0.05. The first stage specifies base learning rate $10^{-3}$, 40 warm-up epochs, and effective batch size 64 (batch 8 with eight-step accumulation); the saved implementation scales this base learning rate by effective batch size over 256. Later stages specify nominal learning rates $2\times10^{-5}$ and $10^{-5}$ with two warm-up epochs, batch size four, and 16-step accumulation. This account describes the original recorded training recipe, not a fresh rerun. The original report identifies an NVIDIA RTX 5090 as the training hardware.

\subsection{Model variants and evaluation protocol}\label{sec:protocol}
The retained implementation contains a standard WildMAE constructor with decoder width 512 and depth 8, as well as a V4Strong constructor with width 768 and depth 16. The corresponding retained vanilla MAE constructor has width 512 and depth 8. All use the same ViT-B encoder family. Their existence documents implemented variants; the surviving materials do not establish a complete checkpoint-to-row mapping for the headline comparison. Consequently, Table~\ref{tab:main} is retained as the original report's result, not presented as a newly generated capacity-matched ablation.

The original evaluation reports PSNR (higher is better), SSIM (higher is better), and LPIPS (lower is better). To translate normalized predictions back into image-space values, the retained reconstruction code uses
\begin{equation}
\widehat{\mathbf p}^{\mathrm{RGB}}_i=\widehat{\mathbf p}_i\sqrt{s_i^2+\epsilon}+\mu_i,
\label{eq:denorm}
\end{equation}
where $\mu_i$ and $s_i^2$ are measured on the full corresponding input patch. Thus the image-space scores characterize \emph{input-statistic-assisted masked reconstruction}; they are not a test of RGB recovery when hidden-patch statistics are genuinely unavailable. The preserved baseline and WildMAE evaluator implementations also differ in aggregation and composite-image construction. We therefore distinguish historical reported numbers from results that would require a newly harmonized common evaluator.

\subsection{Scope of available records}
The present manuscript uses the original course report, surviving source code, manifests, and retained evaluation summaries. Existing saved robustness CSVs permit arithmetic re-aggregation, but image files and model checkpoints were not present in the research package used for manuscript preparation. No new training or checkpoint-based inference was conducted for this manuscript. In particular, there are no newly calculated multi-seed confidence intervals or downstream wildlife-recognition results.

\section{Results}
\subsection{Overall reconstruction observations}
Table~\ref{tab:main} reproduces the primary comparison from the original report on the 4,744-image test split at nominal 75\% masking. Its recorded arithmetic differences are $+3.57$ dB PSNR, $+0.060$ SSIM, and $-0.160$ LPIPS for WildMAE relative to vanilla MAE. The reported values are consistent in direction across pixel, structural, and perceptual metrics, and provide the principal historical observation motivating the proposed refinement design. Because the original checkpoint identities and common evaluator configuration cannot now be established, these differences are descriptive rather than an independently validated controlled estimate of the added modules' effect.
\begin{table}[htbp]
\centering
\caption{\textbf{Original reported reconstruction results} at nominal 75\% masking on 4,744 test images. Historical course-report values; not newly reproduced under a common evaluator.}
\label{tab:main}
\begin{tabular}{lccc}\toprule
Model & PSNR $\uparrow$ & SSIM $\uparrow$ & LPIPS $\downarrow$\\\midrule
Vanilla MAE & 24.33 & 0.780 & 0.320\\
WildMAE & \textbf{27.90} & \textbf{0.840} & \textbf{0.160}\\\bottomrule
\end{tabular}
\end{table}

\subsection{Component variants}
Table~\ref{tab:ablation} preserves the original project's component comparison. The single local-refinement and cross-attention variants each show favorable reported scores relative to the original baseline row. Under the basic pixel loss, the dual-refinement row remains above baseline but does not consistently improve over both individual variants. The strongest reported row combines the dual-refinement model with the staged objective. This pattern motivates examining complementary structure, context, and supervision, but should not be taken as a capacity-matched causal decomposition: complete initialization, model-size, and evaluator provenance for these historical rows is not retained.
\begin{table}[htbp]
\centering
\caption{\textbf{Component-variant results as reported} in the original project at 75\% masking. The rows are descriptive historical records, not a newly verified matched-control series.}
\label{tab:ablation}
\small
\begin{tabular}{@{}p{0.63\linewidth}rrr@{}}\toprule
Variant & PSNR $\uparrow$ & SSIM $\uparrow$ & LPIPS $\downarrow$\\\midrule
Vanilla MAE + basic pixel loss & 24.33 & 0.780 & 0.320\\
Cross-attention refinement only + basic loss & 25.90 & 0.813 & 0.217\\
Local structural refinement only + basic loss & 25.73 & 0.812 & 0.224\\
WildMAE dual refinement + basic loss & 25.75 & 0.810 & 0.221\\
WildMAE dual refinement + staged loss & \textbf{27.90} & \textbf{0.840} & \textbf{0.160}\\\bottomrule
\end{tabular}
\end{table}

\subsection{Synthetic-degradation observations}
The original project also evaluated clean inputs and four synthetic perturbations---Gaussian noise, low light, motion blur, and random occlusion---under 75\% masking. Table~\ref{tab:robust} reproduces rounded mean differences from the retained robustness records; their arithmetic aggregation was independently checked during manuscript preparation. The recorded means favor WildMAE in each condition, with the smallest PSNR separation under low light. These are \emph{differences of saved evaluation means}, not fully controlled paired-corruption estimates. Only 2,372 of 4,744 image records per condition share matching mask hashes across the two model outputs, and identical corrupted inputs cannot be established from the retained files.
\begin{table}[htbp]
\centering
\caption{\textbf{Descriptive differences in retained degradation records} at nominal 75\% masking. $\Delta$PSNR and $\Delta$SSIM are WildMAE minus baseline; LPIPS reduction is baseline minus WildMAE. The original report rounded these quantities to three decimal places.}
\label{tab:robust}
\begin{tabular}{lrrr}\toprule
Input condition & $\Delta$PSNR $\uparrow$ & $\Delta$SSIM $\uparrow$ & LPIPS reduction $\uparrow$\\\midrule
Clean & +2.591 & +0.075 & 0.095\\
Gaussian noise & +1.128 & +0.031 & 0.052\\
Low light & +0.078 & +0.033 & 0.018\\
Motion blur & +1.537 & +0.026 & 0.033\\
Random occlusion & +0.228 & +0.030 & 0.035\\\bottomrule
\end{tabular}
\end{table}

The clean row of Table~\ref{tab:robust} is not interchangeable with Table~\ref{tab:main}: the robustness pipeline computes its masked-region structural and perceptual scores differently from the primary reconstruction scripts. Separate historical experiments with other model versions, masking ratios, or loss schedules are therefore not substituted into the primary comparison.

\section{Discussion and Limitations}
WildMAE's methodological contribution is the combination of a preserved global Transformer decoding pathway with an explicitly local residual branch and a directed masked-query refinement. Zero-initialized gates allow the added branches to enter optimization gradually. The original results provide encouraging, complementary pixel- and perceptual-metric observations under the studied reconstruction setting, while the component table illustrates the relevance of studying the architecture and supervision jointly. These observations do not establish improved downstream encoder representations or a new state of the art.

\paragraph{Evaluation considerations.} The retained RGB reconstruction converts normalized predictions using full-input patch statistics, including those of masked patches. Accordingly, the reported image-quality scores describe statistic-assisted reconstruction, not recovery of fully unavailable image content. Moreover, the available baseline and WildMAE evaluators differ in metric aggregation and masking/composite conventions, while some model variants have different decoder capacities. The archived degradation means are not based on fully matched masks and corruption realizations. Finally, 2,590 of the 4,744 test records share capture-sequence identifiers with training records and all test records share location identifiers; the original image-level split is not a held-out-site assessment. These limitations qualify the strength and generality of the historical comparisons without changing the measurements originally reported.

A focused next step would use a shared evaluator and deterministic image-indexed masks, compare capacity-matched decoders, and test recovery without hidden-patch statistics. Sequence-disjoint evaluation would further clarify generalization to independent camera-trap capture events. Such studies, as well as repeated-seed stability tests, computational cost measurements, and downstream classification or detection, remain future work rather than implied results of this manuscript.

\section{Conclusion}
We presented WildMAE, a structure-aware masked-autoencoding design that preserves global MAE decoding while adding zero-gated local refinement, masked-to-visible cross-attention, and progressive pixel--edge--perceptual supervision. The original Snapshot Serengeti report and retained experiment records describe favorable reconstruction-metric observations under heavy masking and several synthetic degradations. By specifying the implementation and evaluation scope, this manuscript provides a clear account of the proposed approach and a basis for further controlled validation. Future work should assess the method using a harmonized, leakage-free RGB recovery protocol and stronger generalization splits before extending its claims to real-world wildlife-image restoration.

\section*{Research provenance and availability}
This manuscript develops a four-author final project for CIE6032, \emph{Deep Learning Foundations and its Applications}, at The Chinese University of Hong Kong, Shenzhen. It retains original reported measurements and adds manuscript-level clarification and static review of surviving materials; no new training or checkpoint-based inference was performed. A public repository of original project materials is available at \url{https://github.com/Keke946/WildMAE-Wildlife-Image-Reconstruction}; that repository does not supply the original image files or model checkpoints. Any code reuse must respect its third-party notices and applicable source licenses.

\section*{AI assistance disclosure}
Generative AI assisted with drafting, editorial restructuring, literature lookup, static code review, and organization of evidence. It is not an author. All listed human authors must review and take responsibility for the factual claims, permissions, and final manuscript before any public submission.

\end{document}